\documentclass[runningheads]{llncs}

\usepackage{eccv}

\usepackage{eccvabbrv}
\usepackage{algorithm}
\usepackage{algpseudocode}
\usepackage{amsmath}  
\usepackage{xcolor}
\usepackage{graphicx}
\usepackage{booktabs}
\usepackage{booktabs}
\usepackage{multirow}
\usepackage{colortbl}
\usepackage{makecell}
\usepackage{subcaption}
\usepackage{subcaption}
\usepackage{booktabs}
\usepackage{multirow}
\usepackage{makecell}
\usepackage{xcolor}
\usepackage{wrapfig}
\usepackage{booktabs}
\usepackage{makecell}
\usepackage{xcolor}

\usepackage[accsupp]{axessibility}  % Improves PDF readability for those with disabilities.

\usepackage{hyperref}

\usepackage{orcidlink}

\begin{document}

\title{Empowering Vision-Language-Action Model with Memory via Dual-Level Recurrent Queries} 

\titlerunning{ReMem-VLA}

\author{Hang Li\inst{1} \and
Fengyi Shen\inst{1,2} \and
Dong Chen\inst{2} \and 
Liudi Yang\inst{3} \and
Xudong Wang\inst{5} \and \\
Jinkui Shi\inst{5} \and
Zhenshan Bing\inst{4,1} \and
Ziyuan Liu\inst{2} \thanks{Corresponding author.} and
Alois Knoll\inst{1}}

\authorrunning{Li et al.}

\institute{Technical University of Munich, Munich, Germany \and
Huawei Heisenberg Research Center, Munich, Germany \and
University of Freiburg, Freiburg, Germany \and
Nanjing University, China \and
Huawei Technologies, China \\
\email{hang1.li@tum.de, ziyuan.liu1@huawei.com}
}

\maketitle

\begin{abstract}
Vision-language-action (VLA) models for closed-loop robot control are typically cast under the Markov assumption, making them prone to errors on tasks requiring historical context. To incorporate memory, existing VLAs either retrieve from a memory bank, which can be misled by distractors, or extend the frame window, whose fixed horizon still limits long-term retention. In this paper, we introduce ReMem-VLA, a \textbf{\textit{Re}}current \textbf{\textit{Mem}}ory VLA model equipped with two sets of learnable queries: frame-level recurrent memory queries for propagating information across consecutive frames to support short-term memory, and chunk-level recurrent memory queries for carrying context across temporal chunks for long-term memory. These queries are trained end-to-end to aggregate and maintain relevant context over time, implicitly guiding the model's decisions without additional training or inference cost. Furthermore, to enhance visual memory, we introduce Past Observation Prediction as an auxiliary training objective. Through extensive memory-centric simulation and real-world robot experiments, we demonstrate that ReMem-VLA exhibits strong memory capabilities across multiple dimensions, including spatial, sequential, episodic, temporal, and visual memory. ReMem-VLA significantly outperforms memory-free VLA baselines $\pi$0.5 and OpenVLA-OFT and surpasses MemoryVLA on memory-dependent tasks by a large margin.
\end{abstract}

\section{Introduction}
\begin{figure}[tb]
  \centering
  \includegraphics[height=7cm]{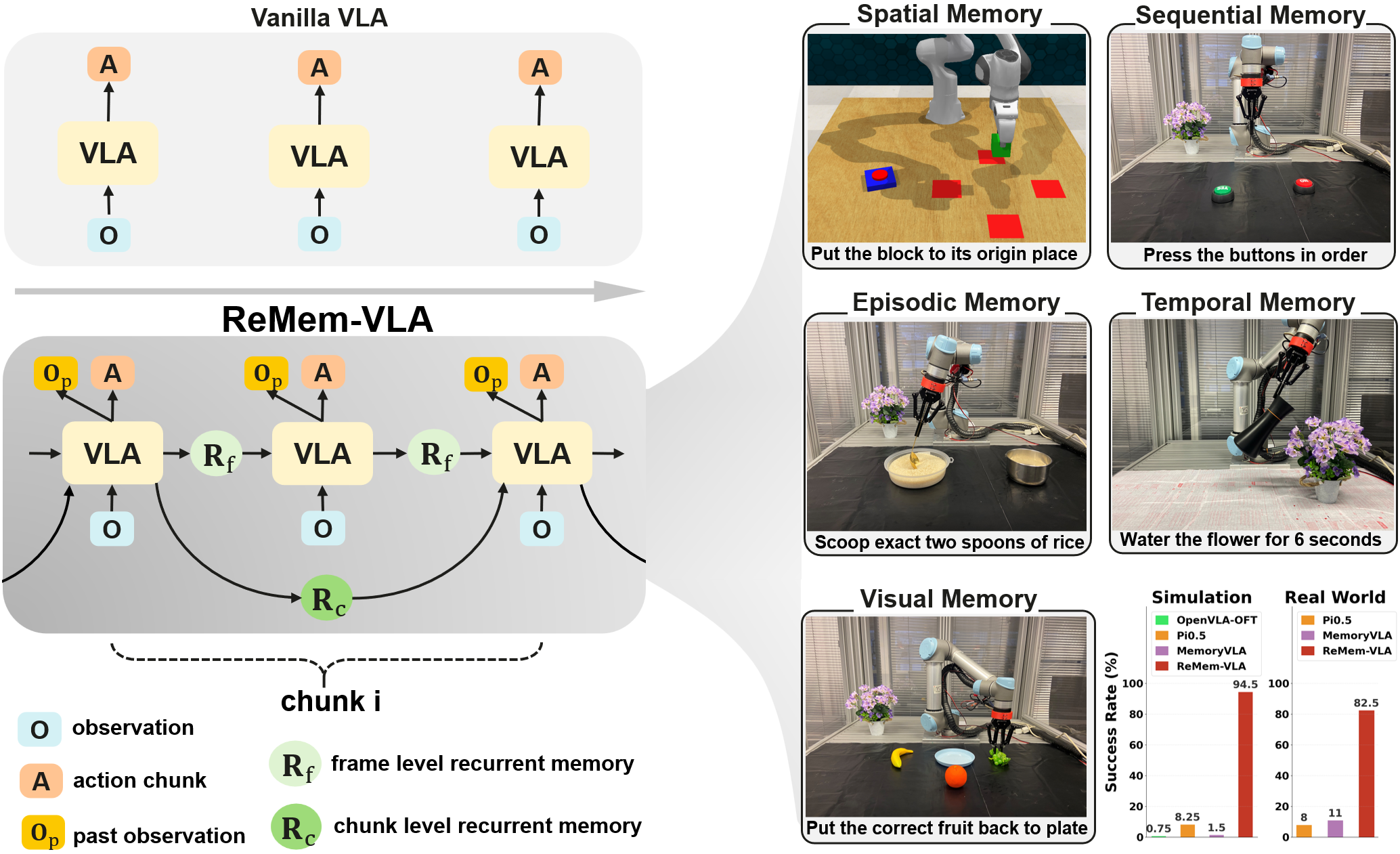}
  \caption{Overview of ReMem-VLA. Compared to vanilla VLA models, ReMem-VLA is equipped with frame-level recurrent memory for short-term retention (e.g., maintaining a fixed pose over several seconds) and chunk-level recurrent memory for long-term memory (e.g., tracking overall task progress). Additionally, ReMem-VLA incorporates past observation prediction to recall historical visual information to strengthen visual memory. ReMem-VLA demonstrates superior performance across spatial, temporal, episodic, sequential, and visual memory benchmarks, substantially outperforming all baselines.}
  \label{fig:overview}
\end{figure}

\label{sec:intro}
Vision-Language-Action (VLA) models \cite{RT-2} have become a central research focus. By leveraging pre-trained vision-language models (VLM) \cite{Llava, Prismatic, PaliGemma, qwen3-vl}, recent VLAs have achieved promising performance on manipulation tasks and demonstrated strong generalization capabilities \cite{Octo, rdt-1b, OpenVLA, InstructVLA,SpatialVLA, UniVLA}. VLAs typically condition on current or short term historical observations \cite{ContextVLA, CronusVLA, HiF} along with language instructions to predict future action chunks, enabling closed-loop control. However, this Markovian paradigm neglects temporal causality, leaving robots blind to what they have done and how the environment has shifted, which leads to systematic failures on memory-dependent manipulation tasks. For example, if a robot is asked to place a cup back to its original location, the VLA has to recall where the cup was originally located. An autonomous robotic system should not only interpret the current state, but also, like humans, leverage cues from the past to derive correct actions for the future. As illustrated in Fig. \ref{fig:overview}, we outline several essential memory capabilities for VLA that remain underexplored: (1) sequential memory, executing multi-step tasks in the correct order; (2) visual memory, using past visual cues to inform current decisions; (3) spatial memory, retaining object locations; (4) episodic memory, tracking previously executed actions; and (5) temporal memory, representing durations and elapsed time. These memories span both short-term retention, such as holding a fixed pose for several seconds, and long-term memory, such as tracking overall task progress. In this work, we target memory for VLA closed-loop robot control, i.e., memory retained within a single task episode across short- and long-horizon tasks, excluding lifelong (persistent) memory that spans sessions over hours, days, or longer timescales.

Recent efforts have augmented VLA or visuomotor policy \cite{ACT, Diffusion-Policy, 3D-Diffusion-Policy, o3dp} with memory through various strategies: history retrieval from memory banks \cite{MemoryVLA, MemER}, extended history horizons \cite{CronusVLA, HAMLET}, sparse history representations \cite{TraceVLA, HistRISE, Bpp} of the complete past episode. However, none of these methods provides a unified memory mechanism that covers the full spectrum of capabilities mentioned above, and each comes with its own limitations: retrieval-based methods relies heavily on the presence of strong cues and suffer from interference at scale; horizon-extension methods fail on tasks exceeding the context window; and point-based or keyframes based memory oversimplify environmental states and is bottlenecked by the reliability of external foundation models \cite{SAM-2, CoTracker, TAPIP3D}. Although \cite{RoboFlamingo, AVA-VLA} introduce RNN-style recurrence into VLAs, they both did not demonstrate long-term memory capabilities. As we show in \ref{Q2}, simple recurrence alone does not yield effective long-term memory in VLAs, because short-window truncated BPTT~\cite{TBPTT} limits temporal credit assignment to only a few steps, preventing the policy from learning dependencies over long horizons.

% To address these challenges, we propose a novel VLA architecture in which the model learns to continuously extract and accumulate implicit task-relevant memory across timesteps, conditioning each step’s prediction on both the current observation and the retained historical context. Specifically, we introduce learnable memory queries into the VLM backbone to extract information from the current observation. At each timestep, the enriched queries are fused with the original queries and fed back as input to the next frame, forming a recurrent loop that continuously propagates critical context across frames without increasing sequence length. However, frame-level propagation alone provides limited memory horizons. To this end, we introduce an additional set of learnable menmory queries that operates across action chunks rather than consecutive frames, evolving at a lower update frequency and retaining information over longer periods. Both query sets are updated using fixed, gradient-free recurrent rules \cref{sec:Gradient-Free Updates}, rather than learnable recurrent dynamics. This dual-level memory design brings substantial benefits to the VLA in capturing both short and long-term contexts.

To address the aforementioned challenges, we propose ReMem-VLA, a novel VLA architecture that facilitates the continuous extraction and accumulation of implicit task-relevant memory. Central to our approach is the integration of learnable recurrent memory queries to VLA model, which enables the model to condition each prediction on a synergized representation of current observations and historical context. To capture multi-scale temporal causality, we architect a dual-level recurrent query scheme: 1) a set of \textbf{frame-level recurrent memory queries} that capture the current frame's context via the VLM and are fused into the next frame's queries in place, forming a recurrent loop that propagates short-term memory across frames without increasing sequence length; and 2) a set of \textbf{chunk-level recurrent memory queries} that update at a coarser temporal granularity across longer temporal intervals, inspired by \cite{Dilated-RNN, Skip-RNN} — unlike frame-level queries which update every step and tend to overwrite earlier context, chunk-level queries accumulate information over longer intervals, enabling stable long-term memory retention beyond what frame-level recurrence can keep. Both recurrent query sets are updated via fixed, gradient-free recurrent path \cref{sec:Gradient-Free Updates} rather than learnable recurrent dynamics. This design allows recurrent queries to learn what task-relevant information to extract and store as memory, while how memory is propagated is fixed and not learned. This design empowers the VLA to effectively associate both short and long-term contexts, enhancing VLA performance on memory-dependent manipulation tasks. Moreover, We further introduce Past Observation Prediction to strengthens visual retention by reconstructing prior observations. 

Concretely, The VLM backbone jointly processes visual observations and language instructions with four types of learnable queries: (1) frame-level recurrent queries for short-term memory; (2) chunk-level recurrent queries for long-term memory. (3) action queries condition to an action diffusion module for predicting the future action chunk; (4) hindsight queries condition to an image head for past RGB image reconstruction; To allow our dual-level recurrent queries to implicitly influence action prediction and visual past recovery, we introduce a bidirectional attention transformer as a connector between the VLM backbone and the prediction heads, where all four types of queries interact in latent space through self-attention. Through end-to-end training with only action and past image prediction objectives, the recurrent queries learn to adaptively extract task required memory without additional supervision.

Training recurrent models requires processing each episode strictly in temporal order without cross-episode state leakage, which is challenging to batch efficiently when episodes have variable lengths. We propose a new training paradigm — slot-based streaming training — where instead of slicing episodes into fixed chunks like conventional approaches, each slot continuously tracks a live episode and contributes one frame per training step, allowing recurrent state to accumulate naturally across the full episode. 

The effectiveness of ReMem-VLA is validated through both simulation and real-world experiments. In simulation, we evaluate on MemoryBench~\cite{SAM2ACT} and introduce a long-horizon task to assess spatial and long-term sequential memory. Ablation studies confirm that dual-level recurrent queries are critical for memory retention, and that fixed recurrent memory propagation path ~\cref{sec:Gradient-Free Updates} rather than learnable path are essential for long-term memory under the TBPTT~\cite{TBPTT} training paradigm. In real-world experiments, we verify the memory capabilities of ReMem-VLA across multiple memory dimensions — episodic, visual, sequential, and temporal — and further confirm the contribution of the Past Observation Prediction.

In summary, our contributions are as follows:

\begin{itemize}
\item We propose ReMem-VLA, a recurrent memory-augmented VLA model featuring dual-level recurrent queries operating at two different temporal granularities, endowing the model with both short-term and long-term memory for memory-dependent manipulation tasks.
\item  We introduce a training paradigm that enables batch training of recurrent VLA over variable-length episodes using independent per-episode slots, without breaking temporal continuity or mixing information across episodes.
\item We introduce a novel past visual observation prediction strategy to strengthen visual memory retention in VLA models.
\item We validate ReMem-VLA through extensive simulation and real-world experiments across spatial, temporal, sequential, episodic, and visual memory tasks, obtaining consistent and substantial performance gains over baselines.
\end{itemize}

\section{Related Work}

\subsection{Vision Language Action Models}

Vision-Language-Action models leverage pre-trained vision-language models to perform robotic manipulation tasks. Existing VLAs adopt two primary strategies for action generation: (1) Discrete action approaches like RT-2 \cite{RT-2}, OpenVLA \cite{OpenVLA} and FAST \cite{fast} tokenize actions or treat actions simply as text, exemplified by VLM2VLA \cite{VLM2VLA} and VLA-0 \cite{vla-0}, enabling autoregressive training with minimal modifications to the VLM backbone. (2) Continuous action approaches, represented by $\pi$0 \cite{pi0}, $\pi$0.5 \cite{pi0.5}, OpenVLA-OFT \cite{OpenVLA-OFT}, GR00T-N1 \cite{GR00T} and InternVLA-M1 \cite{Internvla-m1}, introduce additional action heads on top of the VLM, trained with regression \cite{OpenVLA-OFT, VLM4VLA}, diffusion \cite{GR00T, Internvla-m1}, or flow matching losses \cite{pi0, pi0.5} to output continuous action trajectories. Continuous action formulations generally provide smoother trajectories and better support for real-time control. Following this paradigm, we adopt an action diffusion module to predict continuous action chunks conditioned on VLM features. To improve manipulation performance and generalization, most VLA models undergo pre-training on large-scale robot datasets, e.g., Open-X-Embodiment \cite{OXE}, DROID \cite{driod}, Bridge \cite{bridge}, and AgiBot World \cite{agibot-world}. Notably, recent work VLA-Adapter \cite{vla-adapter} has demonstrated that competitive performance can be achieved without expensive large-scale robot pre-training by introducing learnable action queries that serve as an interface to bridge VLM representations and policy outputs. Similarly, Our VLA model employs learnable action queries to extract features from freezed VLM Backbone and condition the action diffusion module, without large-scale pre-training. An emerging trend in VLA research combines action prediction with explicit future state prediction, demonstrated by GR-1 \cite{GR-1}, Seer \cite{Seer}, DreamVLA\cite{DreamVLA}, and InternVLA-A1 \cite{Internvla-a1}, or video generation approaches \cite{CoVAR, Motus}, such as Genie-Envisioner \cite{Genie-Envisioner}, mimic-video \cite{mimic-video} and Cosmos Policy \cite{Cosmos-Policy}. Predicting images provides fine-grained visual guidance for action generation beyond the high-level semantic understanding from VLM conditioning. We focus on memory capability and, for the first time, augment the VLA with an additional image head to predict \textit{past} RGB images for visual memory enhancement. While these methods advance VLA capabilities in various aspects, they all condition on a single frame or a short window of recent observations, adopting a Markovian framework that cannot leverage extended historical context, which severely hinders performance on memory-dependent manipulation tasks. Our work focuses on enabling VLA to perceive, retain, and leverage complete historical context through recurrent memory mechanisms, addressing memory-dependent tasks that current approaches fail to handle.

\subsection{Visuomotor Policies with History Awareness}
Prior approaches to incorporating historical memory into visuomotor policies can be categorized into several paradigms: (1) Extending observation windows. CronusVLA \cite{CronusVLA} concatenates multi-frame perceptual features to condition action diffusion models, HAMLET \cite{HAMLET} aggregates moment tokens over a fixed 8-step window, and PTP \cite{PTP} learns long-context policies by predicting past tokens during training to capture temporal dependencies. Although these methods modestly improve short-term memory, fixed-length windows cannot adapt to varying memory requirements—history beyond the window is discarded. PAM \cite{PAM} addresses state ambiguity by maintaining a working memory of up to 300 frames through adaptive memory recoding, yet tasks exceeding this fixed horizon remain unaddressed. (2) Sparse history representations. TraceVLA \cite{TraceVLA} overlays past frames as point traces on current observations to encode motion history, and HistRISE \cite{HistRISE} leverages object-centric point tracking as additional input. Bpp \cite{Bpp}learns policies by conditioning on a few VLM-selected keyframes from the history. Though capable of processing full episode history, sparse history representations omit essential details and depend heavily on external model stability. (3) Memory bank retrieval. MemoryVLA \cite{MemoryVLA} introduces a memory bank similar to agentic framework, training the model to retrieve relevant historical tokens. MemER \cite{MemER} extends this paradigm by constructing a keyframe-based experience bank for retrieval at inference time. But the retrieval quality is query-dependent, degrading when current frames lack strong cues or when similar distractors introduce interference. (4) Recurrent mechanisms. RoboFlamingo \cite{RoboFlamingo} employs an LSTM-style policy head with recurrent hidden states, and AVA-VLA \cite{AVA-VLA} maintains a recurrent belief state to modulate visual attention. However, neither work evaluates the memory capabilities of their models on memory-dependent tasks, they both evaluated only on standard manipulation tasks. Critically, as we demonstrate in our ablation studies (\ref{Q2}), naive recurrence trained via truncated backpropagation \cite{TBPTT} fails to provide effective memory, particularly over longer horizons. We validate that fixed gradient-free memory propagation and introducing chunk-level recurrence prove essential for robust memory. We introduce dual recurrent query streams—frame-level for short-term and chunk-level for long-term dependencies—that adaptively capture task-relevant context across arbitrary horizons.

\section{Method}
\subsection{Problem Formulation}
Consider a multi-task manipulation dataset $\mathcal{D} = \{(\ell_i, \{o_t, a_t\}_{t=0}^{T_i})\}_{i=0}^{N}$ consisting of $N$ tasks. Each task $i$ contains number of $T_i$ demonstrations. And each demonstration is defined by a language instruction $\ell_i$ and $T_i$ observation-action pairs $(o_t, a_t)$. Here, the observations $o_t \in \mathbb{R}^{V \times h \times w \times 3}$ correspond to RGB camera images from $V$ camera views, while the actions $a_t \in \mathbb{R}^n$ represent the control commands.
A typical VLA model $\pi_\theta$ follows a Markovian decision-making formulation, predicting future action chunks based solely on the current observation and language instruction(and possibly proprioceptive inputs):
\begin{equation}
\mathcal{A}_{t:t+k} = \pi_\theta(o_t, \ell),
\end{equation}
where $\mathcal{A}_{t:t+k} = (a_t, \ldots, a_{t+k})$ represents a 
sequence of $k$ future actions (action chunk size k). We extend the VLA model to a history-aware formulation with auxiliary past observation prediction:
\begin{equation}
\mathcal{A}_{t:t+k}, \hat{o}_{t-m} = \pi_\theta(o_t, \ell, h_t)
\end{equation}
where $h_t$ is a recurrent memory state that accumulates historical context over time.
Specifically, $h_t = (h^c_t, h^f_t)$ consists of two components operating at different
granularities: a frame-level memory updated at every timestep,

\begin{equation}
h^f_{t} = \mathcal{F}_f(h^f_{t-1},\ o_{t-1}, \ell)
\label{eq:frame_memory}
\end{equation}
and a chunk-level memory updated at chunk boundaries,
\begin{equation}
h^c_{t} = \mathcal{F}_c(h^c_{t-k},\ o_{t-k}, \ell, h^f_{t-k})
\label{eq:chunk_memory}
\end{equation}
where $\mathcal{F}_f$ and $\mathcal{F}_c$ denotes the memory update path.

\subsection{Model Architecture}
\begin{figure}[tb]
  \centering
  \includegraphics[height=5.5cm]{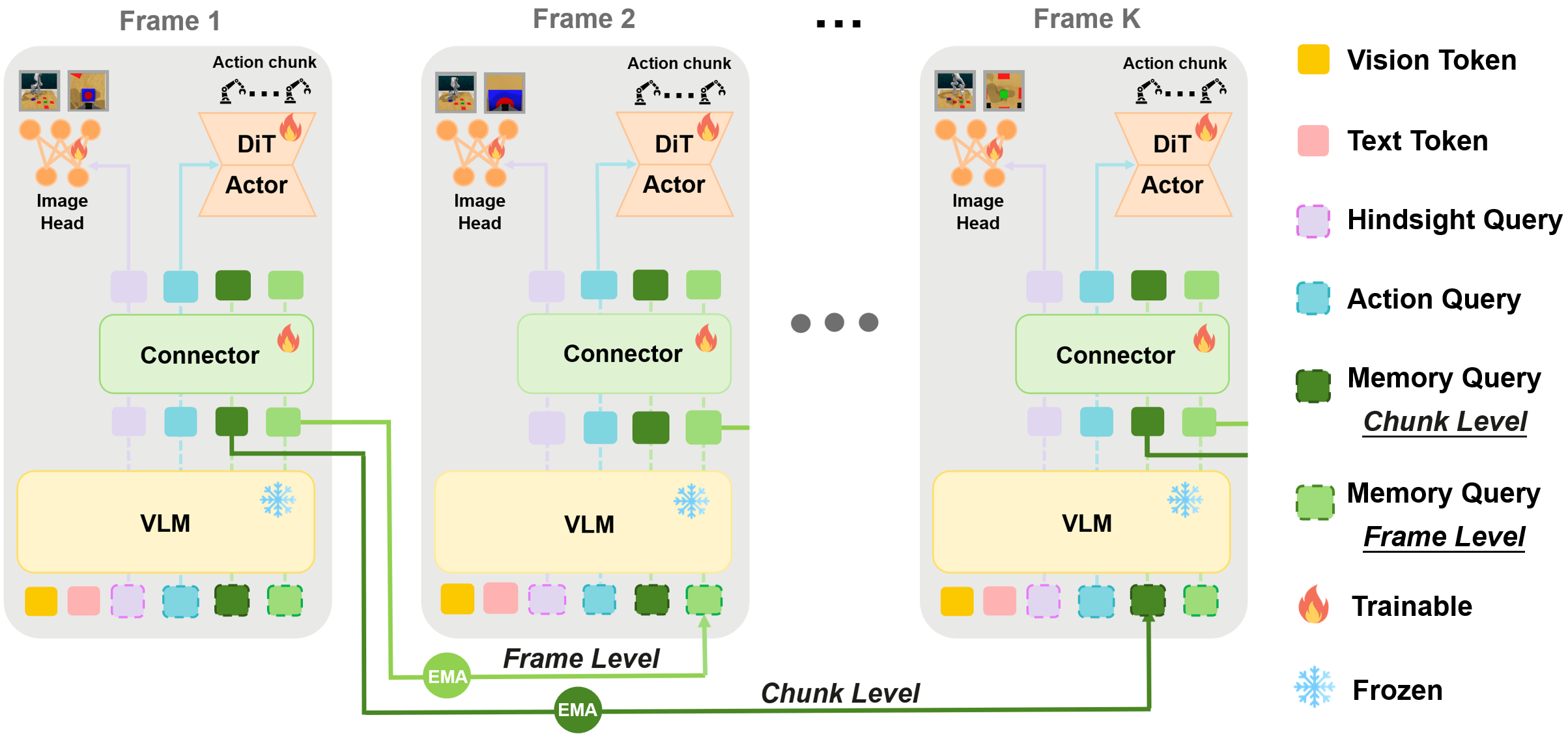}
  \caption{Model Architecture of ReMem-VLA. ReMem-VLA continuously processes visual observations and language instructions at each timestep, predicting an action chunk and a past frame. Simultaneously, two sets of learnable recurrent memory queries extract information from the VLM backbone and propagate it to future timesteps via exponential moving average (EMA) updates. Among them, Frame-level memory queries are updated at every frame for short-term memory, while chunk-level memory queries are updated only at chunk boundaries (every $K$ frames) for long-term retention. Action queries and hindsight queries extract features from the VLM and interact with these recurrent memory queries through bidirectional attention in connector, fusing current observations with accumulated historical context to condition the predictions.}
  \label{fig:model}
\end{figure}

% 细节太多，删减不必要的点
We design ReMem-VLA as an end-to-end framework for robotic manipulation, as shown in \cref{fig:model}. The architecture comprises five key components: (1) a frozen VLM backbone that processes multimodal inputs—RGB observations $o_t \in \mathbb{R}^{V \times H \times W \times 3}$ and language instructions $\ell$; (2) learnable action queries $Q^{action}$ and hindsight queries $Q^{img}$ that extract features from the VLM for action and image prediction; (3) dual-level recurrent memory queries—frame-level $Q^f$ and chunk-level $Q^c$—that accumulate historical context across timesteps; (4) a transformer-based connector with bidirectional attention enabling interaction among all queries; and (5) prediction heads for future action generation and past image reconstruction.

\subsubsection{VLM Backbone and Task-Specific Learnable Queries}
We adopt Qwen3-VL-2B \cite{qwen3-vl}, a 2-billion parameter vision-language model, as our frozen backbone \cite{vla-adapter}. At each timestep, the backbone processes RGB observations through its vision encoder and language instructions through its text encoder, producing hidden states $\mathbf{H}_t \in \mathbb{R}^{L \times D}$, where $L$ is the sequence length and $D$ is the hidden dimension. The backbone maintains causal attention internally. We introduce two sets of task-specific learnable queries appended to the input sequence: action queries $Q^{action} \in \mathbb{R}^{N_a \times D}$ and hindsight queries $Q^{img} \in \mathbb{R}^{N_{img} \times D}$ \cite{GR-1}, where $N_a$ and $N_{img}$ denote the number of query tokens. $Q^{action}$ focuses on extracting features critical for action generation, capturing manipulation-relevant semantics, while $Q^{img}$ extracts features for predicting past visual observation, recalling previously seen views.

\subsubsection{Dual-Level Recurrent Memory Queries}
\label{sec:Gradient-Free Updates}
To enable memory capabilities across different temporal scales, we introduce dual-level recurrent queries: frame-level memory queries $Q^f \in \mathbb{R}^{N_f \times D}$ for short-term retention and chunk-level memory queries $Q^c \in \mathbb{R}^{N_c \times D}$ for long-term retention. At each timestep, both sets of memory queries attend to VLM hidden states and the action and hindsight queries ($Q^{action}$, $Q^{img}$) to extract current-frame features, yielding updated query representations $\tilde{Q}^f_t$ and $\tilde{Q}^c_t$ that encode the observed information at timestep $t$. These representations are then propagated to future timesteps via exponential moving average~(EMA).

\textbf{Frame-level memory} updates densely at every timestep:
\begin{equation}
Q^f_t = \beta_f \cdot \tilde{Q}^f_{t-1} + (1-\beta_f) \cdot Q^f_{t-1}
\end{equation}
where $\beta_f \in [0,1]$ is a fixed EMA coefficient. This dense frame-by-frame 
propagation provides fine-grained short-term memory for tracking recent object 
states, ongoing actions, and immediate temporal dependencies.

\textbf{Chunk-level memory} updates sparsely at chunk boundaries (every $k$ frames):
\begin{equation}
Q^c_t =
\begin{cases}
\beta_c \cdot \tilde{Q}^c_{t-k} + (1-\beta_c) \cdot Q^c_{t-k}, 
& \text{if } t \bmod k = 0 \\[6pt]
Q^c_{t-1}, 
& \text{otherwise}
\end{cases}
\end{equation}
where $\beta_c \in [0,1]$ is the chunk-level EMA coefficient. The lower update frequency results in slower exponential decay $(1-\beta_c)^{n/k}$ compared to frame-level $(1-\beta_f)^n$, enabling stable long-horizon memory that retains initial object configurations, task progress, and long temporal dependencies.

%definition
\textbf{Gradient free recurrent update path $\mathcal{F}$.} 
In our architecture, the recurrent update path $\mathcal{F}$ (equation \ref{eq:frame_memory} and \ref{eq:chunk_memory}) consists of the VLM backbone and the EMA-based fusion mechanism. Ideally, training a recurrent model to capture long-term dependencies would require backpropagating gradients through the entire trajectory. However, for large VLA models, this is computationally infeasible. In practice the training relies on Truncated BPTT~\cite{TBPTT} with a small truncation window (e.g. T=4 in AVA-VLA\cite{AVA-VLA}), which truncates the computation graph so gradients do not propagate past the window boundary; therefore the action loss cannot teach the model what to retain over hundreds of timesteps. To remove this bottleneck, we make the recurrence path $\mathcal{F}$ gradient-free (frozen VLM + fixed EMA), so memory propagation is guaranteed by a deterministic forward update; learning is only about what information the memory queries write, not how memory is propagated. Ablations in Q2 \ref{Q2} further show that introducing learnable recurrent coefficients or unfreezing the VLM substantially degrades long-term memory performance.

\subsubsection{Connector}
Since the VLM maintains causal attention, $Q^{action}$ and $Q^{img}$ cannot attend to memory queries within the backbone. We introduce a transformer connector \cite{metaquery} between the VLM and prediction heads. The connector is an 12-layer transformer with bidirectional self-attention, where all four query types—$Q^{action}$, $Q^{img}$, $Q^f$, and $Q^c$—interact comprehensively. This allows $Q^{action}$ and $Q^{img}$ to integrate their current-frame features with the accumulated temporal context in $Q^f$ and $Q^c$, producing memory-enriched representations for downstream predictions.

\subsubsection{Action and Image Heads}
We adopt a diffusion model to predict action chunks $\mathcal{A}_{t:t+k} \in \mathbb{R}^{k \times 7}$ conditioned on action queries 
$Q^{action}$ via cross-attention. Following DDPM \cite{ddpm}, we train the denoising network with:
\begin{equation}
\mathcal{L}_{\text{action}} = \mathbb{E}_{\tau, \epsilon \sim \mathcal{N}(0,I)} 
\left[ \| \epsilon - \epsilon_\theta(\mathcal{A}_\tau, \tau, Q^{action}) \|^2 \right],
\end{equation}
where $\mathcal{A}_\tau$ is the noised action at diffusion timestep $\tau$, 
and $\epsilon_\theta$ predicts the added noise. During inference, we use DDIM \cite{ddim} for faster sampling. We employ a lightweight ViT-style \cite{ViT} patch decoder conditioned on hindsight queries $Q^{img}$ to reconstruct the past observation $o_{t-m}$, supervised by MSE loss:
\begin{equation}
\mathcal{L}_{\text{image}} = \| o_{t-m} - \hat{o}_{t-m} \|_2^2.
\end{equation}
The overall training objective combines both losses:
\begin{equation}
\mathcal{L}_{\text{total}} = \mathcal{L}_{\text{action}} + 
\lambda_{\text{img}} \mathcal{L}_{\text{image}}.
\end{equation}

\subsection{Training and Inference Recipe}

\textbf{Streaming slot-based batching for recurrent training.}
Training recurrent robot policies over episodes presents a batching challenge: episodes vary in length, and the model must process frames strictly in temporal order without cross-episode contamination. Chunk-based batching cuts episodes into fixed-length windows for easy batching, but it breaks temporal continuity and discards long-horizon context.
We instead adopt a slot-based streaming approach. A fixed pool of $\mathcal{B}$ concurrent slots is maintained, each tracking a different episode at its current timestep. At every training step, one frame is sampled from each slot in sequence, producing a batch of size $\mathcal{B}$ without any padding or truncation. Within a slot, the recurrent state persists and accumulates context across the full episode duration. At episode boundaries, we hard-reset the slot’s recurrent state, guaranteeing strict isolation and preventing any cross-episode state leakage.

\textbf{Inference.}
At each timestep, two sets of recurrent query tokens are maintained and updated continuously: frame-level queries refreshed at every step, and chunk-level queries updated at a coarser frequency. Both are reset only when a new episode begins. On a single RTX 4090 GPU, our model has inference frequency of ~10 Hz. If actions are predicted less frequently and the last predicted action chunk is reused for intermediate control steps, the effective control rate can be increased to nearly 20 Hz. Details are provided in the appendix.

\section{Experiments}
\subsection{Implementation Details}
The visual input to ReMem-VLA are a third view and a wrist view, both at 256×256 resolution. We use absolute joint positions and binary gripper state as the action space, with an action chunk size of 30. Robot proprioceptive state is optionally as an additional input. We set the image loss weight $\lambda_{\text{img}}$ to 0.5. During inference, actions are generated via DDIM with 20 denoising steps. All experiments are conducted on 8xA100 GPUs with a total batch size of 64, and inference is deployed on a single GPU. We adopt a cosine learning rate schedule with an initial learning rate of 5e-5 and a minimum learning rate of 1e-7. We set BPTT truncation horizon to 1 during training.

\subsection{Simulation Experiments}

\begin{figure}[tb]
  \centering
  \includegraphics[height=3.3cm]{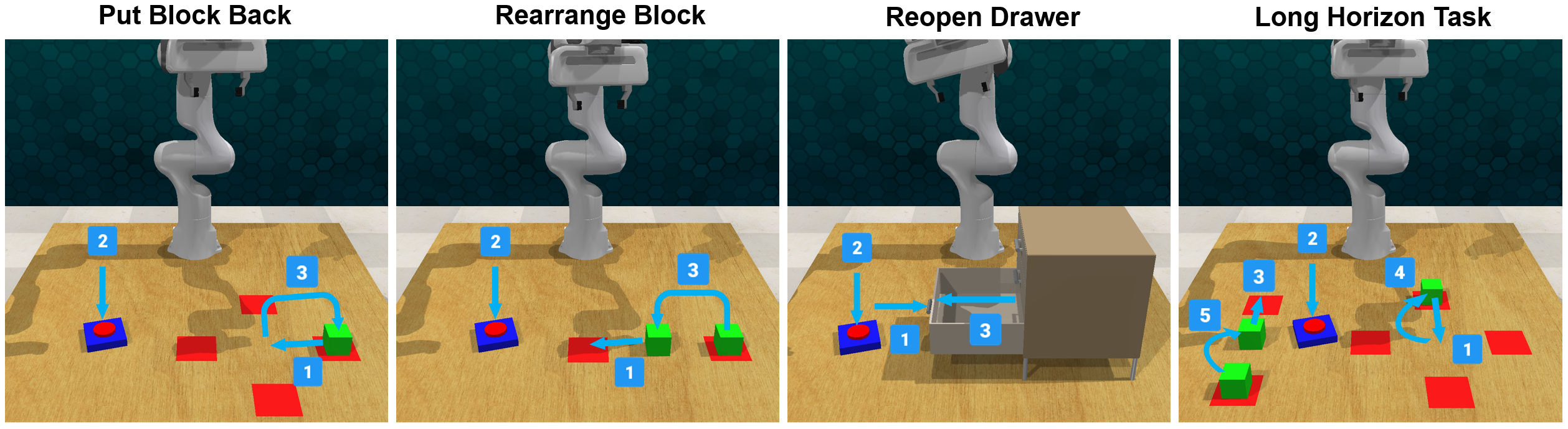}
  \caption{MemoryBench \cite{SAM2ACT} with our extended long horizon task. 1)Put Block Back: Place the block to the center, presses a button, and then returns the block to its original location. 2)Rearrange Block: Move the block from the center to the unoccupied red patch, presses a button, and then relocates the block that was originally on the red patch to the center. 3)Reopen Drawer: Close the open drawer, presses a button, and then reopen the same drawer. 4) Long Horizon Task: We extend the memorybench with an additional long horizon task to evaluate the VLA's memory over long horizon. This task is a combination of Put Block Back and Rearrange Block.}
  \label{fig:sim}
\end{figure}

The experiments focus on evaluating the memory capabilities of VLA models. We therefore adopt MemoryBench \cite{SAM2ACT} as our simulation benchmark, which comprises three memory-dependent tasks, see Fig. \ref{fig:sim}: 1) Put Block Back. 2) Rearrange Block. 3) Reopen Drawer. These three tasks collectively require spatial memory, and logical reasoning over history. However, as each task spans approximately only 300 frames, we further introduce an extended 4) Long Horizon Task exceeding 600 frames that demands memory retention, to more rigorously assess VLA's memory capacity on long-horizon tasks. This task is constructed by composing Rearrange Block and Put Block Back, as shown in Fig. \ref{fig:sim}.

% 任务描述太多
\begin{table}[tb]
\centering
\caption{\textbf{The extended MemoryBench \cite{SAM2ACT} experiments.} ReMem-VLA achieves the best performance across all tasks compared to baselines.}
\label{MemoryBench}
\begin{tabular}{l|cccc|c}
\toprule
\multirow{2}{*}{Methods} & \multicolumn{4}{c|}{Scores (\%)} & \multirow{2}{*}{Average} \\
\cmidrule(lr){2-5}
& \makecell{Put \\ Block back} & \makecell{Rearrange \\ Block} & \makecell{Reopen \\ Drawer} & \makecell{Long \\ Horizon Task} & \\  % <-- 关键：补这个 &
\midrule
OpenVLA-OFT ~\cite{OpenVLA-OFT}   & 0 & 0 & 3 & 0 & 0.75 \\
$\pi_{\text{0.5}}$ ~\cite{pi0.5}  & 6 & 4 & 20 & 3 & 8.25 \\
MemoryVLA ~\cite{MemoryVLA}       & 0 & 1 & 5 & 0 & 1.5 \\
\rowcolor{gray!15}
ReMem-VLA (ours)                  & \textbf{93} & \textbf{99} & \textbf{100} & \textbf{86} & \textbf{94.5} \\
\bottomrule
\end{tabular}
\end{table}

\subsubsection{Training, Evaluation Setup, and Results}
For each task, we collect 100 training demonstrations using front and wrist cameras at $256\times256$, train jointly on all four tasks for 150k steps, and evaluate with 100 rollouts per task. We compare ReMem-VLA against standard VLAs (OpenVLA-OFT \cite{OpenVLA-OFT} and $\pi_{\text{0.5}}$ \cite{pi0.5}) and MemoryVLA \cite{MemoryVLA}, a retrieval-based memory VLA, reproducing all baselines under the same data and evaluation protocol. The original MemoryBench \cite{SAM2ACT} setup often fails when actions hit joint limits, obscuring memory-related errors; we reduce button-position randomization to 70$\%$ to avoid such failures. In \textit{Rearrange Block}, different scene configurations yield visually distinct trajectories that enable cue-based solving, so we enforce consistent trajectories across configurations to make memory necessary. Table \ref{MemoryBench} shows ReMem-VLA reaches 94.5$\%$ average success over four tasks, surpassing all baselines.

\subsection{Real World Experiments}
\subsubsection{Real world tasks design}
\begin{figure}[tb]
  \centering
  \includegraphics[height=7.0cm]{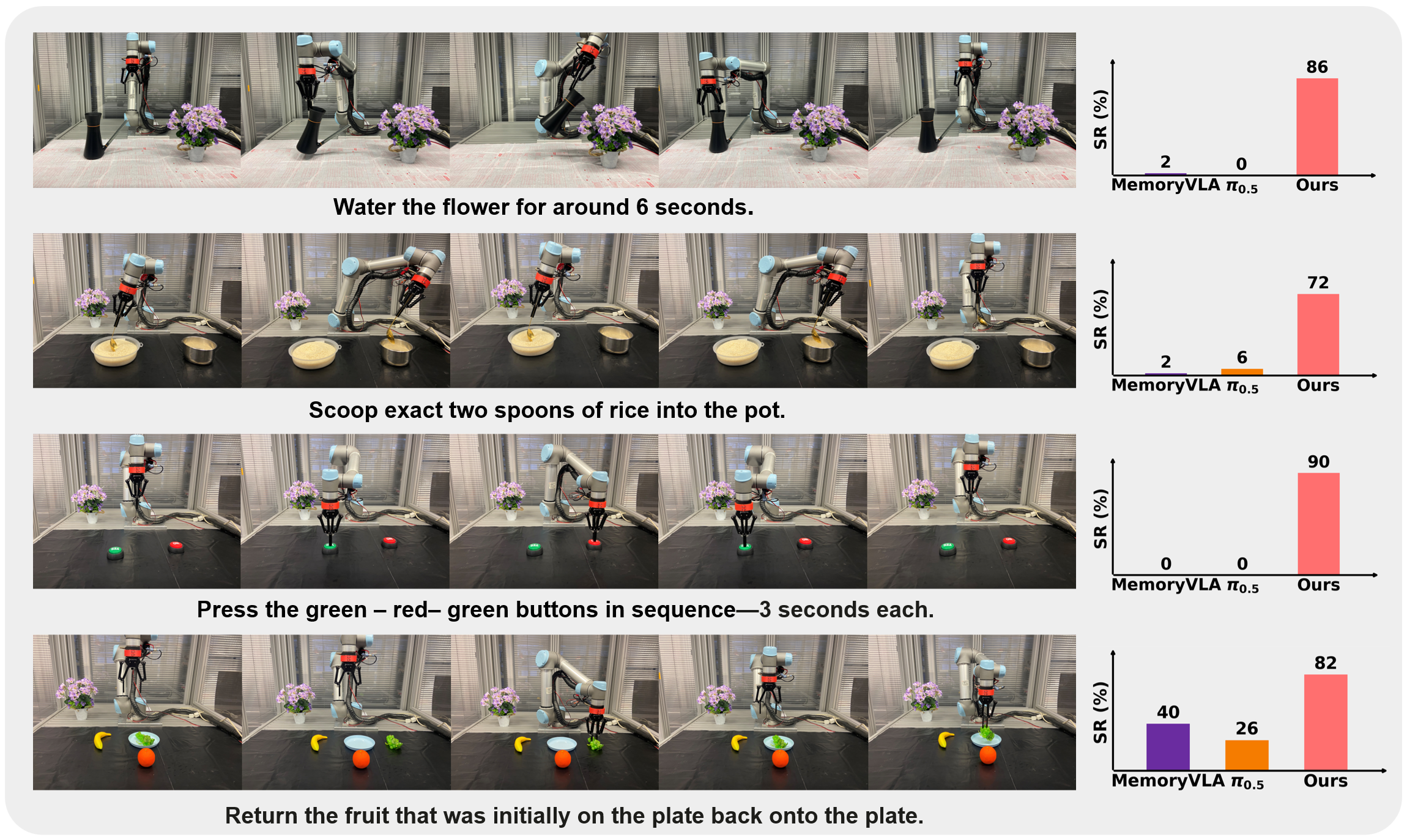}
  \caption{Real world experiments and quantitative results on memory-dependent tasks.}
  \label{fig:real}
\end{figure}

We design four real-world manipulation tasks that require memory to evaluate model performance in real-world scenarios (Fig. \ref{fig:real}). Since MemoryBench \cite{SAM2ACT} only assesses spatial memory, lacking evaluation of temporal memory, sequential memory, episodic memory and visual memory, we design four tasks to probe each capability respectively. (1) \textbf{Water flower for around six seconds}: this task lacks salient visual feedback, the model must maintain the watering pose for approximately six seconds, reflecting its ability of temporal memory.  (2) \textbf{Scoop two spoons of rice into pot}: the robot must scoop exact two spoons of rice into the pot and then stop, requiring episodic memory. (3) \textbf{Press buttons in sequence, and each for 3 seconds}: the robot must press greed-red-green buttons in order and press each for around 3 seconds. This task requires both memory of sequential ordering and temporal memory. (4) \textbf{Put fruit back to the plate}: the robot must return the fruit that was originally on the plate back to it, requiring the model to remember the initial visual scene, evaluating visual memory capability.

\subsubsection{Training, Evaluation Setup, and Results}
Real-world experiments are conducted on a UR5 robot arm with a Robotiq gripper. Both wrist and third-person views are captured using RealSense D435 cameras. For each task, 200 demonstrations are collected via SpaceMouse teleoperation, with object positions randomized within a predefined range. Each task is evaluated over 50 trials and reported as success rate. We use $\pi_{\text{0.5}}$ ~\cite{pi0.5} and MemoryVLA \cite{MemoryVLA} as real-world baseline. For the Water Flower and Press Button tasks, duration measurement is only approximate within an range due to the human collected demonstrations. In the Press Button task, we occasionally introduce disturbances by suddenly shifting the button position to verify true closed-loop control and prevent the task from being solved by an open-loop extreme long trajectory. ReMem-VLA achieve an average success rate of 82.5$\%$, far outperforming MemoryVLA \cite{MemoryVLA} and $\pi_{0.5}$ \cite{pi0.5} (8$\%$ and 11$\%$, respectively), demonstrating comprehensive memory capabilities across visual spatial, temporal, and episodic aspects.

\subsection{Ablation Study}
In this section, we conduct ablation studies to answer the following questions.

\begin{figure}[tb]
  \centering
  \begin{subfigure}[t]{0.49\linewidth}
    \centering
    \includegraphics[height=4.0cm]{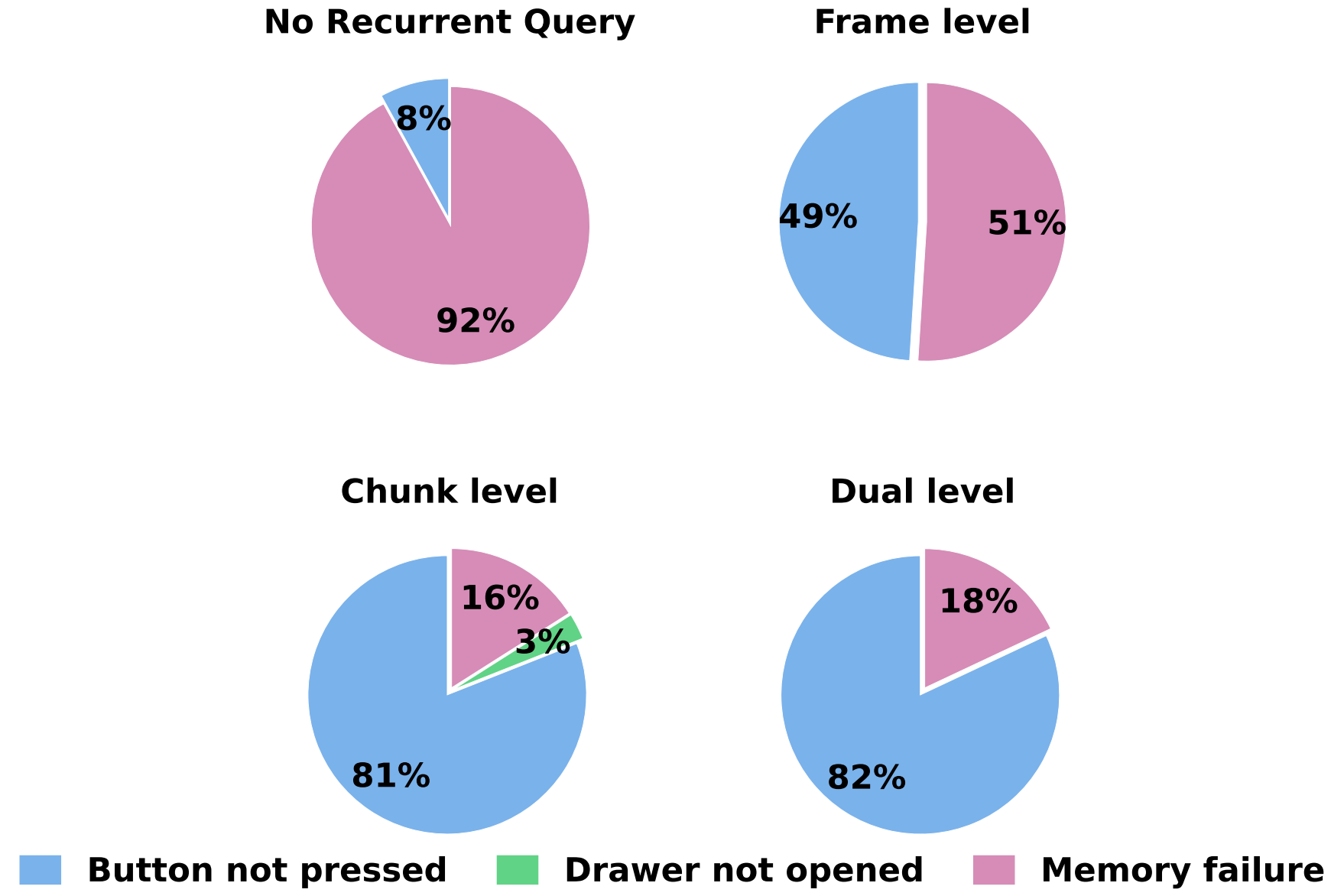}
    \caption{Failure analysis in simulation.}
    \label{fig:failure}
  \end{subfigure}
  \hfill
  \begin{subfigure}[t]{0.49\linewidth}
    \centering
    \includegraphics[height=4.0cm]{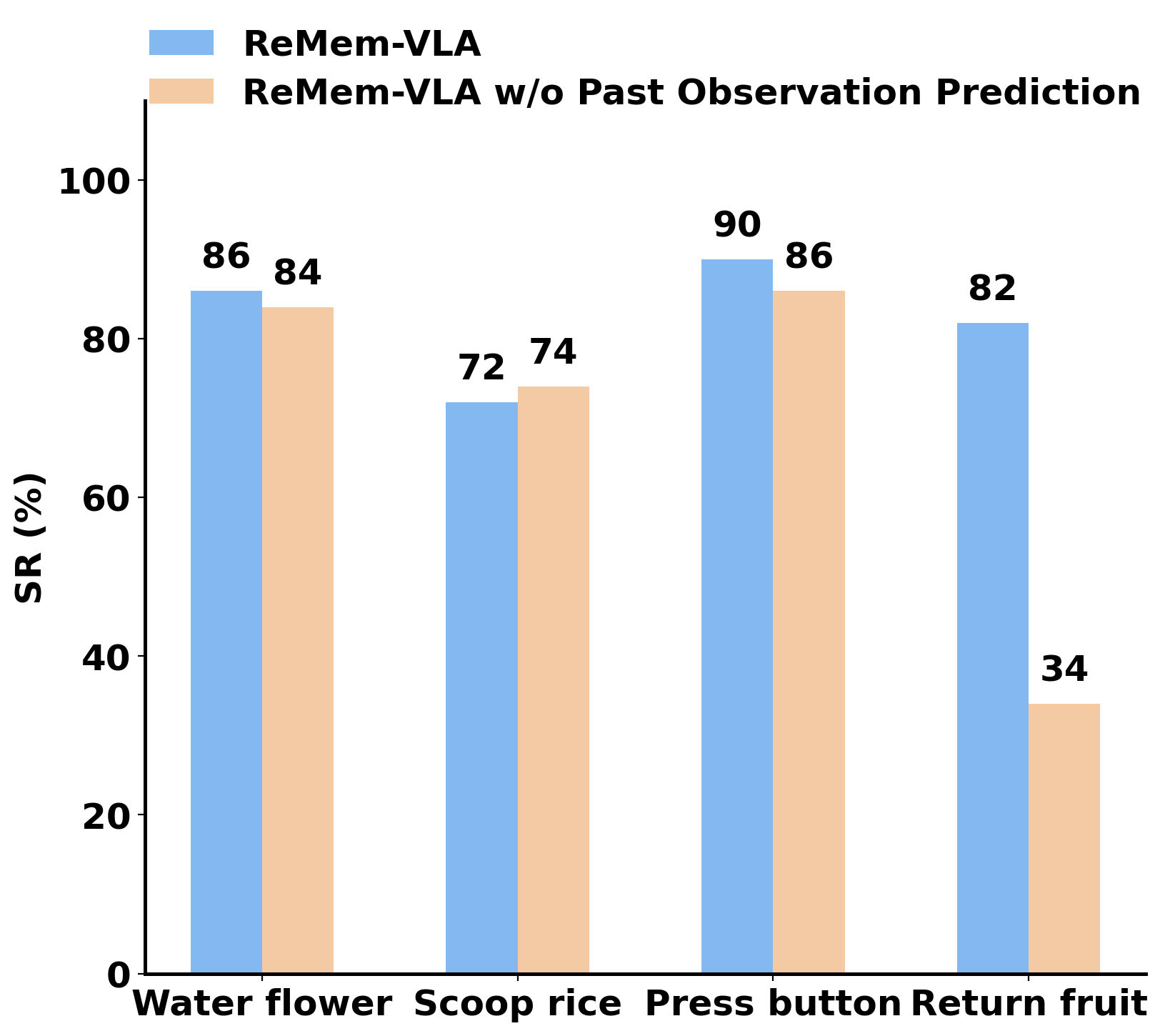}
    \caption{Ablation on POP in real world.}
    \label{fig:pop}
  \end{subfigure}
  \caption{Failure analysis and Ablation on Past Observation Prediction(POP).}
  \label{fig:failure_pop}
\end{figure}

\textbf{What is the contribution of each level of recurrent queries to VLA's memory capabilities?} To investigate the contribution of each level of recurrent queries to the memory capability of VLAs, we compare four configurations: (1) no recurrent queries, (2) frame-level queries only, (3) chunk-level queries only, and (4) dual-level queries combining both frame level and chunk level. To avoid confounding effects from differing numbers of learnable queries, we fix the total query count to 128 across all settings: the dual-level model allocates 64 to each level. We evaluate all configurations on the four MemoryBench tasks (Table \ref{tab:query_levels}) and further analyze the distribution of failure causes (Fig. \ref{fig:failure}). We find that removing all recurrent queries causes a complete collapse of memory capability, with success rates dropping dramatically across all tasks. Among the single-level variants, frame-level queries alone struggle with long-term memory, leading to a higher proportion of memory-related failures, while chunk-level queries alone improve long-term retention but result in a lower overall success rate, as pressing the button demands short-term memory that chunk-level queries cannot provide. The dual-level model achieves consistently better performance across all tasks, demonstrating that both levels of recurrent queries contribute complementary memory capabilities to the model.

\begin{table}[t]
\centering
\caption{\textbf{Ablation on different recurrent-query levels.}}
\label{tab:query_levels}
\begin{tabular}{l cccc c}
\toprule
\multirow{2}{*}{Methods} & \multicolumn{4}{c}{Scores (\%)} & \multirow{2}{*}{Average} \\
\cmidrule(lr){2-5}
& \makecell{Put \\ Block back} & \makecell{Rearrange \\ Block} & \makecell{Reopen \\ Drawer} & \makecell{Long \\ Horizon Task} & \\
\midrule
No Recurrent Query & 7  & 32 & 27 & 5  & 17.75 \\
Frame Level        & 90 & 95 & 96 & 70 & 87.75 \\
Chunk Level        & 81 & 92 & 94 & 71 & 84.5  \\
\rowcolor{gray!15}
Dual Level         & \textbf{93} & \textbf{99} & \textbf{100} & \textbf{86} & \textbf{94.5} \\
\bottomrule
\end{tabular}
\end{table}

\textbf{The effectiveness of Past Observation Prediction.}
To validate Past Observation Prediction (POP), we conduct an ablation study on the four \textit{real-world tasks}. As shown in Fig. \ref{fig:pop}, Past Observation Prediction provides limited benefit on tasks that do not rely on visual memory, but substantially improves the visually memory-intensive Return Fruit task (34$\%$ $\rightarrow$ 82$\%$), especially when we set to predict the episode’s first frame. Our results also show that, while dual-level recurrent queries yield strong memory in spatial, temporal, episodic, and sequential aspects, they are less effective at preserving visual details and benefit from Past Observation Prediction.

\begin{figure}[tb]
  \centering
  \includegraphics[height=2.6cm]{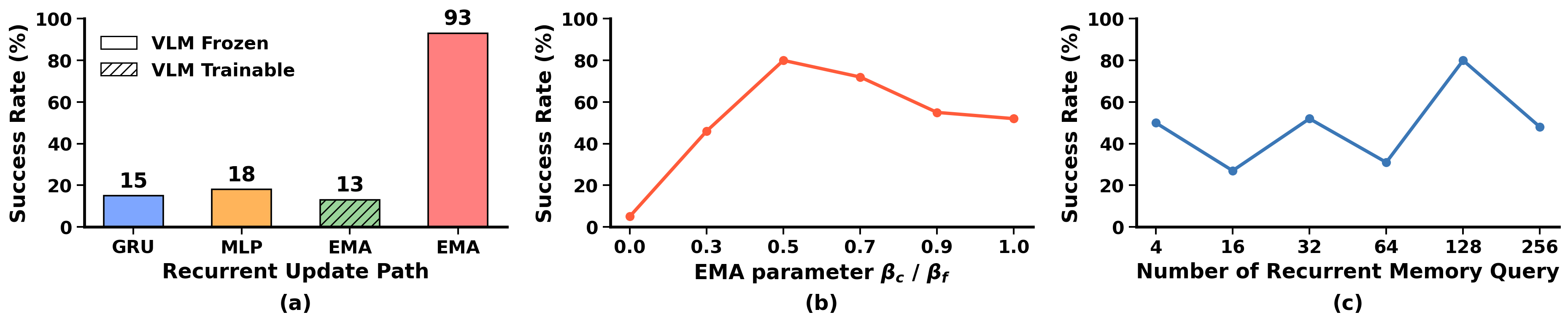}
  \caption{Ablation study on (a) different recurrent path; (b) number of recurrent queries. (c) different EMA parameters }
  \label{fig:ablation}
\end{figure}

\textbf{How does Gradient-free recurrent update path $\mathcal{F}$ compare to trainable recurrent dynamic in terms of performance?} To validate the effectiveness of our gradient-free recurrent update path $\mathcal{F}$  (equation \ref{eq:chunk_memory},\ref{eq:frame_memory}), we compare a frozen vs. trainable VLM backbone, and replace the EMA update with a learnable GRU or a simple MLP on the \textit{put block back} simulation task. As shown in Fig. \ref{fig:ablation} (a), We find that introducing trainable parameters into the recurrent update path (trainable VLM / GRU / MLP) almost completely eliminates the model’s memory capability.
\label{Q2}

\textbf{How do the EMA hyperparameters $\beta_c$ and $\beta_f$ affect performance?}
We sweep the EMA coefficient ($\beta_f = \beta_c\in\{0,0.3,0.5,0.7,0.9,1\}$) on the \textit{long horizon task} task in simulation (Fig.~\ref{fig:ablation} (b)) and observe the best performance at $\beta=0.5$. We attribute this to the retention--adaptation trade-off in EMA updates: larger $\beta$ updates memory too aggressively and quickly overwrites useful history, while smaller $\beta$ makes memory too inert to absorb new task-relevant information.

\textbf{Effect of the number of recurrent memory queries.}
We vary the number of recurrent memory queries ($N\in\{4,16,32,64,128,256,512\}$) and observe that a moderate $N$ (128) performs best on the \textit{Long Horizon Task} in simulation (Fig. \ref{fig:ablation} (c)). With too few queries, memory capacity is insufficient to preserve task-relevant context, while too many queries introduce redundancy that can increase attention noise and make optimization harder.

\textbf{Impact of the chunk-level recurrent update interval.} We set the chunk-level update interval to {0.5, 1, 2, 3}$\times$ the action chunk size (30). This interval balances stability and freshness of chunk-level memory, avoiding stale updates or excessive overwriting. Across MemoryBench tasks, $1\times$ yields the best overall performance. Experimental details and results are provided in the appendix.

\section{Conclusion}
This paper proposes ReMem-VLA, a VLA model with a dual-level recurrent memory mechanism. The frame-level recurrence provides short-term memory, while the chunk-level recurrence supports long-horizon retention. We further introduce Past Observation Prediction to improve visual memory. We enable efficient batch training on variable-length episodes by resetting recurrent states at episode boundaries. Across memory-demanding manipulation tasks in both simulation and real-robot settings, ReMem-VLA demonstrates strong memory capability, covering spatial, temporal, visual, episodic, and sequential aspects. Despite these advantages, a current limitation is that our model has not been trained on large-scale robot datasets, which may affect its generalization ability. Future work includes pretraining ReMem-VLA on large datasets, or integrating our memory mechanism into existing pretrained VLA models.

\bibliographystyle{splncs04}
\bibliography{main}
\end{document}